\documentclass{bmvc2k}

\title{~~~~InteriorNet: Mega-scale Multi-sensor \textcolor{white}{=--}Photo-realistic Indoor Scenes Dataset}

\addauthor{~~~~~~Wenbin Li}{wenbin.li@imperial.ac.uk}{1}
\addauthor{~~~~~~Sajad Saeedi}{s.saeedi@imperial.ac.uk}{1}
\addauthor{~~~~~~John McCormac}{brendan.mccormac13@imperial.ac.uk}{1}
\addauthor{~~~~~~Ronald Clark}{ronald.clark@imperial.ac.uk}{1}
\addauthor{~~~~~~Dimos Tzoumanikas}{dimosthenis.tzoumanikas14@imperial.ac.uk}{1}
\addauthor{~~~~~~Qing Ye}{zhentou@qunhemail.com}{2}
\addauthor{~~~~~~Yuzhong Huang}{yuzhongh@usc.edu}{2}
\addauthor{~~~~~~Rui Tang}{ati@qunhemail.com}{2}
\addauthor{~~~~~~Stefan Leutenegger}{s.leutenegger@imperial.ac.uk}{1}

\addinstitution{
Department of Computing\\
Imperial College London\\
London UK, SW7 2AZ
}
\addinstitution{
KooLab, Kujiale.com\\
Hangzhou China
}

\runninghead{Li \etal}{InteriorNet: Mega-scale Photo-realistic Dataset}

\def\etal{\emph{et al}\bmvaOneDot}

\begin{document}
\maketitle

\begin{abstract}
Datasets have gained an enormous amount of popularity in the computer vision community, from training and evaluation of Deep Learning-based methods to benchmarking Simultaneous Localization and Mapping (SLAM). Without a doubt, synthetic imagery bears a vast potential due to scalability in terms of amounts of data obtainable without tedious manual ground truth annotations or measurements. Here, we present a dataset with the aim of providing a higher degree of photo-realism, larger scale, more variability as well as serving a wider range of purposes compared to existing datasets. Our dataset leverages the availability of millions of professional interior designs and millions of production-level furniture and object assets -- all coming with fine geometric details and high-resolution texture. We render high-resolution and high frame-rate video sequences following realistic trajectories while supporting various camera types as well as providing inertial measurements. Together with the release of the dataset, we will make executable program of our interactive simulator software as well as our renderer available at \url{https://interiornetdataset.github.io}. To showcase the usability and uniqueness of our dataset, we show benchmarking results of both sparse and dense SLAM algorithms.
\end{abstract}

 \begin{figure}[h]
 \centering
 \includegraphics[width=0.98\linewidth]{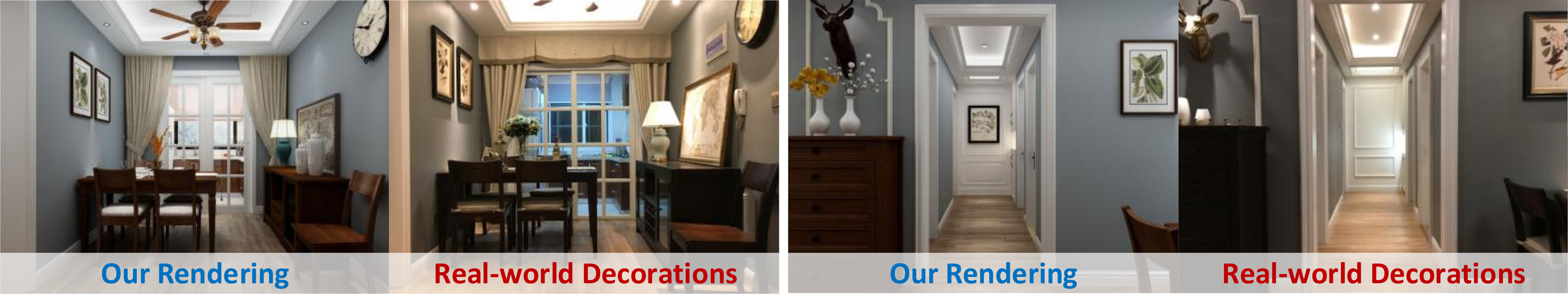}
 \vspace{-2mm}
 \caption{Our rendering vs.\ real decoration guided by our furniture models and layouts.\vspace{-2mm}
 }
 \label{fig:decoration}
 \vspace{-4mm}
 \end{figure}
 
\vspace{-6 mm}
\section{Introduction}
\vspace{-2.5 mm}
Spatial perception has undergone a true step-change during the past few years, partly thanks to the Deep Learning revolution in Computer Vision, but not only. Spatial perception -- core to e.g.\ a robotic system -- involves a plethora or elements that should be addressed jointly: these include the understanding of camera pose relative to the environment, i.e.\ Simultaneous Localization and Mapping (SLAM), which ideally includes a very dense reconstruction of at least the immediate surrounding environment, allowing the robot to interact with it. As purely geometric understanding is still limiting, the research community has been shifting towards tightly integrating it with semantic understanding, object-level mapping, estimating dynamic content, etc.

In order to benchmark SLAM, as well as to both train and evaluate semantic understanding, datasets of increasing scale have been playing a central role. While we argue that real recorded datasets will remain crucial for the foreseeable future, they suffer from scalability limitations: for instance preparing ground truth scene models is a challenging task, with methods such as laser scanning being a costly and time-consuming process. Thanks to advances in computer graphics and computational power, the use of rendered synthetic models is an obvious choice and recent works such as ICL-NUIM dataset~\cite{Handa_ICRA_2014} for SLAM and SceneNet RGB-D dataset~\cite{Mccormac:ICCV:2017} for semantic labeling are examples of such works. In this paper, we have improved upon current rendering methods and datasets first of all by proposing a versatile and fast rendering framework for a high degree of photo-realism which leverages the availability of \textit{millions of realistic indoor designs} composed of \textit{millions of detailed digital object models}. Importantly, we model realistic lighting and scene \textit{change over time}. Aside from RGB rendering, we can opt for depth images and semantics. Second, we use and synthesize realistic trajectories as ground truth to render at \textit{video frame rate} from various classes of typical motion patterns.
Fig.~\ref{fig:decoration} shows samples from a real-world environment and synthesized images generated from models. In summary, our contributions are:

\begin{itemize}\vspace{-2.5 mm}
\item  Our scene database contains around 1M furniture CAD models and 22M interior layouts (Sec.~\ref{sec:overview}), which have been created for real-world production and decoration.
\vspace{-3 mm}
\item Our dataset contains two parts: (1) 15k sequences rendered from 10k randomly selected layouts, with 1k images for each sequence; (2) 5M images are rendered from 1.7M randomly selected layouts, with 3 images per layout.\vspace{-3 mm}
\item To simulate realistic scenes that change over time (Sec.~\ref{sec:scene}), we interfaced a physics engine for object movement and provide flexibility to manipulate lighting conditions. \vspace{-3 mm}
\item We introduce a learned algorithm to realistically style random  trajectories (Sec.~\ref{sec:traj}). \vspace{-3 mm}
\item We implemented a fast, photo-realistic renderer, called \textit{ExaRenderer} (Sec.~\ref{sec:renderer}).\vspace{-3 mm}
\item We implemented a user friendly simulator, called \textit{ViSim} (Sec.~\ref{sec:visim}), to assist creating monocular or stereo camera trajectories and synthesize related ground truth.
\vspace{-3 mm}
\item We show the usefulness of the dataset by providing SLAM evaluation results
(Sec~\ref{sec:evaluation}). \vspace{-3 mm}
\item We will release our dataset (consisting of the rendered sequences and images) as well as \textit{ExaRenderer}, \textit{ViSim}, and a subset of the 3D models and layouts used for evaluations.
\end{itemize}


\begin{figure*}[t]
\centering
\includegraphics[width=.97\linewidth]{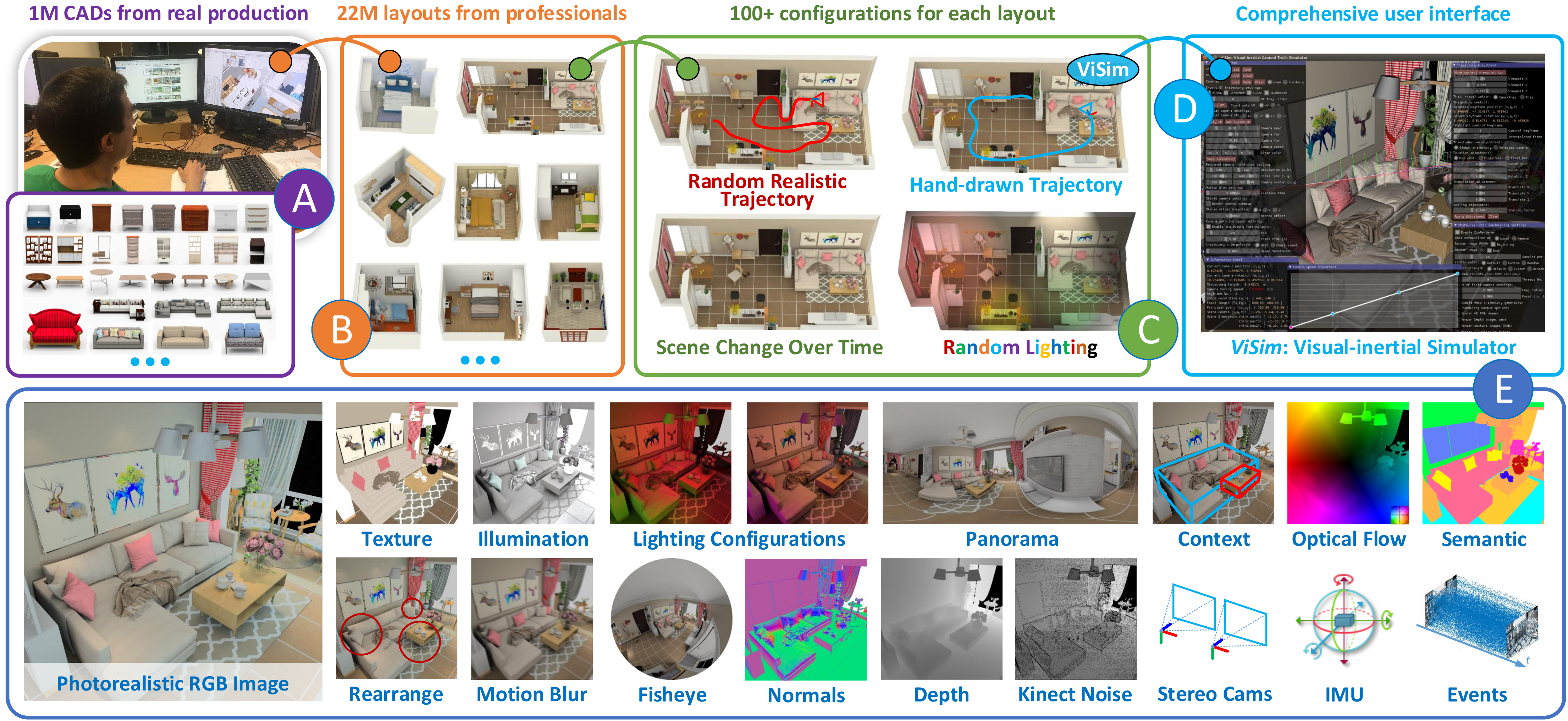}
\vspace{-3mm}
\caption{System Overview: an end-to-end pipeline to render an RGB-D-inertial benchmark for large scale interior scene understanding and mapping. (\textbf{A}) We collect around 1 million CAD furniture models from world-leading furniture manufacturers. These models have been used in the real-world production. (\textbf{B}) Based on those models, around 1,100 professional designers/companies create around 22 million interior layouts. Most of such layouts have been used in real-world decorations. (\textbf{C}) For each layout, we generate a number of configurations to represent different lightings and simulate scene change over time in daily life. (\textbf{D}) We provide an interactive simulator (\textit{ViSim}) to create the ground truth monocular/stereo trajectories, as well as IMU and event camera data. Trajectories can be set manually, or using random walk and neural network based generation. (\textbf{E}) All supported image sequences and ground truth data.}
\label{fig:teaser}
\vspace{-4mm}
\end{figure*}

\vspace{-2 mm}
\section{Background}\label{sec:background}
\vspace{-2 mm}
Benchmarking algorithms with real-world and synthetic datasets for various computer vision tasks has been playing an important role, especially in the field of robotics.
In this section, we review popular datasets used in semantic labeling and SLAM.
In semantic labeling, the two most closely related synthetic datasets are the SUN-CG~\cite{song2016ssc} with photo-realistic rendering~\cite{Zhang:etal:CVPR2017}, and SceneNet RGB-D~\cite{Mccormac:ICCV:2017}.
SUN-CG also provides realistic hand-designed layouts similar to ours but from Planner 5D~\cite{planner5d}, a layout tool for amateur interior design.
The number of layouts (45K), unique object meshes  (2.6K), and rendered images (\cite{Zhang:etal:CVPR2017} 500K) are orders of magnitude smaller and without the daily life noise added, when compared to our proposed dataset -- and importantly, we focus on rendered full trajectories rather than still images.
SceneNet RGB-D~\cite{Mccormac:ICCV:2017} provides full trajectories, but at much lower sample rate and of only one type, whereas here we provide a number of base random trajectories, as well as a learned style for realistic jitter.
SceneNet RGB-D also contains fewer images (5M) which are of a lower resolution (320$\times$240) and quality, containing circular artefacts from the photon mapping process.
The object models in SceneNet RGB-D are from ShapeNet~\cite{Shapenet:ARXIV2015} which contains 51K models from public online repositories.
Although an excellent resource, the quality of the models and textures from these repositories can be quite variable and missing ground truth metric scales requires automated prediction~\cite{Savva:2014:RSS:2670291.2670295,DBLP:conf/cvpr/SavvaCH15}.
Additionally, Zhang \etal~\cite{zhang:arXiv:2016} show that pretraining with a synthetic dataset improves the results of computer vision tasks such as surface normal prediction, semantic segmentation, and object boundary detection.
Qi \etal~\cite{qi2018human} use human context to simulate layout of the indoor environment.
Here we use only assets of commercial production quality, with accurate metric scales and high quality textures at a much larger scale, all semantically labelled by the manufacturer.
The closest related work with real-world data is the ScanNet dataset~\cite{dai2017scannet}.  It is a dataset consisting of 2.5M frames, containing trajectories of 1.5K indoor scenes. The scenes are manually annotated with a tool designed for use in Amazon Mechanical Turk.

There are many real-world visual localization and mapping datasets.
Examples are the TUM RGB-D dataset with 19 different trajectories~\cite{Sturm_IROS_2012}, the EuRoC dataset with 11 visual and inertial sequences~\cite{Burri:IJRR_2016}, the KITTI dataset with large-scale 2D outdoor trajectories~\cite{Geiger_CVPR_2012}, and New College dataset with medium scale trajectories. Ground truth for these trajectories is either obtained from GPS or motion capture systems, and for the map from laser scanners (in KITTI and EuRoC only). In the active vision dataset~\cite{Ammirato_ICRA_2017},  limited control options are provided to control the trajectory in 2D while navigating through the images. These datasets are very useful, but either the number of trajectories or the environments are limited.

Among synthetic visual SLAM datasets,
UnrealCV~\cite{Qiu_ArXive_2016}, \cite{zhang:arXiv:2016}, uses the Unreal Engine to render realistic images for labelling and SLAM.
The ICL-NUIM dataset~\cite{Handa_ICRA_2014} provides eight sequences from two models using POV-Ray renderer~\cite{povray}. 
These methods require manual work to setup the rendering environment and also the number of the models is limited.

\vspace{-2 mm}
\section{Dataset Overview}\label{sec:overview}
\vspace{-2 mm}
The proposed pipeline (Fig.~\ref{fig:teaser}) provides a large scale furniture model database and interior layouts, plus an end-to-end rendering pipeline to create the proposed RGB-D-inertial data.
\vspace{-2 mm}
\subsection{Furniture Models and Interior Layouts}
\vspace{-2mm}

\begin{figure*}[t]
\centering
\includegraphics[width=.97\linewidth]{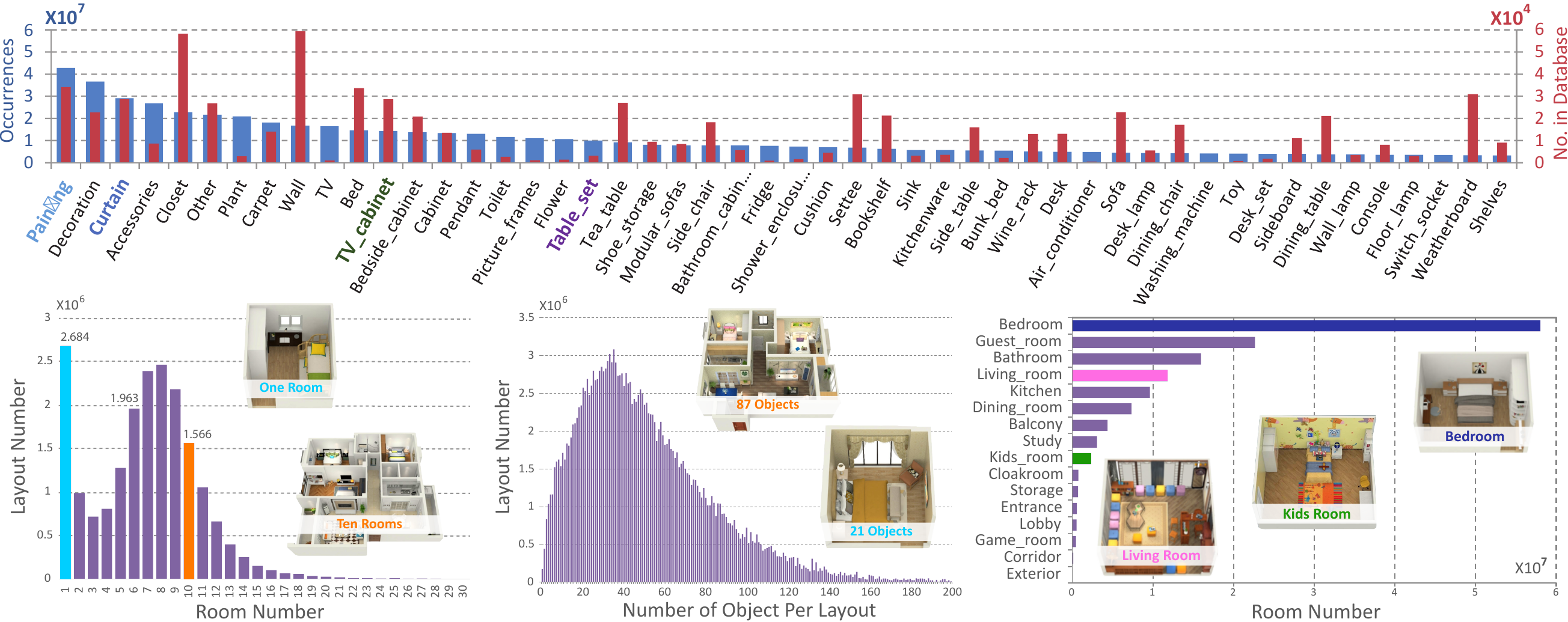}
\vspace{-4mm}
\caption{Statistics on object models, rooms and layouts. (\textbf{Top}): occurrences of the 50 most common categories of objects from the layouts (\textit{blue}), and number of objects in our database (\textit{red}). (\textbf{Bottom Left}): distribution of number of rooms per layout. (\textbf{Bottom Middle}): distribution of number of object per layout. (\textbf{Bottom Right}): Distribution of room type.}
\label{fig:object}
\vspace{-3mm}
\end{figure*}

We collected 1,042,632 computer-aided design (CAD) models of furnitures from 42 world-leading manufacturers. Those models have been categorized into 158 main classes and are 
manually mapped to NYU40 categories \cite{Silberman:ECCV12}. Fig.~\ref{fig:object} shows the statistics and sample images of the top 50 categories in our database. Our furniture model database provides several unique features: (1) all object meshes are measured in real-world dimensions and feature the exact same measurements as their real-world counterparts; (2) all object meshes are high resolution (triangles and vertex) and associated with hierarchical semantic labeling information. For instance, a specific sofa model contains 734,198 triangles, every vertex of which is  labeled in terms of parts i.e.\ arms, legs, seat cushion, back foam and frame; (3) all objects are used in real production and will be easily spotted in real-world homes; (4) every object is associated with multiple textures and materials. For example, every chair has different color schemes, covers and accessories. The material information of the models is provided by manufacturers, which is compatible to \textit{VRayStyle}~\cite{vray}. Material representation is guided by compositing 4 Bidirectional Reflectance Distribution Function (\textit{BRDF}) properties: lambertian, microfacet, dielectric and transmission.

Given such a large-scale object database, 1,078 professional designers working with end customers have created 22,652,123 interior layouts for different scenarios since Oct.\ 2014. Fig.~\ref{fig:object} illustrates the statistics of our layouts which contain 16 types of rooms adopted from the interior design industry. The most representative room types are bedroom, guest room, bathroom, living room and kitchen. Our layouts are diverse in range from studio to 42-room apartments.
Most of the layouts are being used in real-world decoration.
\vspace{-2 mm}
\subsection{Configurations}
\vspace{-2 mm}
For each layout, we generated a number of different configurations to either rearrange the furniture or randomized the lighting conditions (Fig.~\ref{fig:lighting}). For the former (Sec.~\ref{sec:rearrange}), we implemented a physics based automatic framework to slightly rearrange the movable objects. For the latter (Sec.~\ref{sec:sim_lighting}), we support both manual or automatic changes on color and intensity of the default lighting. These setups diversify our layouts to simulate traces of daily life and the changes of natural lightings through a day.

\vspace{-2 mm}
\subsection{Rendered Image Sequences and Ground Truth}
\vspace{-2 mm}
We propose an end-to-end pipeline to render photo-realistic images and associated ground truth data (Fig.~\ref{fig:teaser} (E)). We created multiple trajectories (Sec.~\ref{sec:traj}) for each layout by varying the combination of linear velocity, angular velocity and trajectory types. Each view of a trajectory consists of both a shutter open and shutter close camera pose (many renderings from poses in-between are averaged to obtain motion blurred renderings). Our images support 640$\times$480 resolution, at 25 Hz, resulting in 1,000 images per trajectory. By using an optimal parameter setup for best image quality, each render takes less than 2 ms on our cluster of 1,300 Nvidia GTX TiTanX GPUs and 2,000 Intel Xeon Phi CPUs. Our pipeline supports different RGB variations (pure texture, illumination and normals), and different types of lens models (panorama, fisheye and depth of field). The rendering speed may vary for rendering variations (a 5,000$\times$2,500 panorama image takes 0.5-1 sec on our cluster to render).

\begin{figure*}[t]
\centering
\includegraphics[width=1\linewidth]{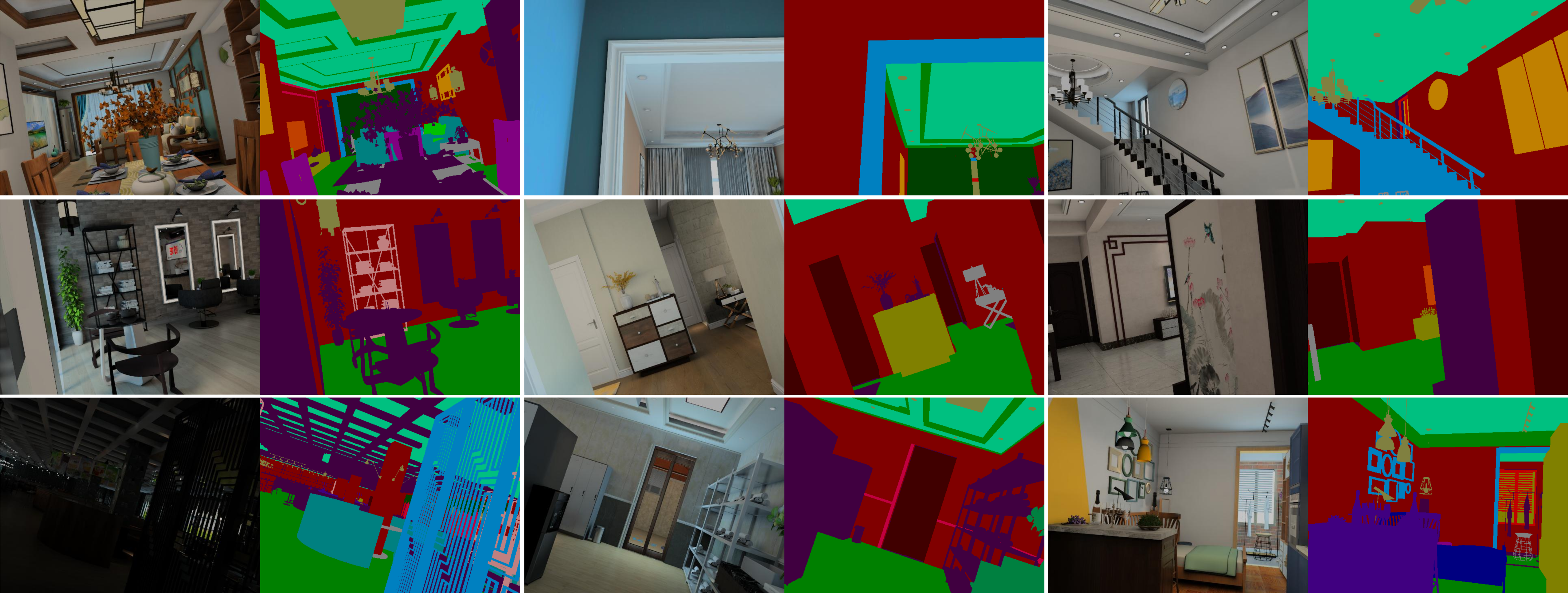}
\vspace{-5mm}
\caption{Our rendered images and the associated NYU40 labels.}
\label{fig:nyu}
\vspace{-6mm}
\end{figure*}

Our rendering pipeline is able to create various per-frame ground truth data. Per-pixel semantic labels e.g. NYU40 (Fig.~\ref{fig:nyu}) can be obtained from model settings (provided by manufacturers) and a rendering pass. Per-object semantic context (3D bounding box) is provided to indicate the 3D extent of an object in the scene. 
Depth is obtained as the Euclidean distance of the ray intersection; and noisy depth is generated by simulating a real \textit{Kinect} mechanism~\cite{choo2014statistical}. We also generate instance segmentations and optical flow following~\cite{Mccormac:ICCV:2017}.

\vspace{-2 mm}
\subsection{Dataset Structure and Hosting}
\vspace{-2 mm}
Our dataset contains 20M photo-realistic images and many forms of ground truth data on around 1.7M layouts randomly selected from our database.
To organize this wealth of data, we separated sequences into 20 different subsets in terms of number of rooms. We also provide a sparse subset that contains a smaller number of images (5M) but at highest diversity. We randomly selected 1.7M layouts, for each of which a random view was rendered in three different configurations, i.e.\ original layout, random lighting and random rearrangement. The parameters for randomly choosing lighting and rearrangement remained the same. The images of each subset mentioned above were randomly divided into 80\%, 10\% and 10\% splits for training, validation and test sets respectively. We also release a subset of CAD models and layouts, as well as the rendering pipeline and the simulator (\textit{ViSim}) for public.

\vspace{-2 mm}
\section{Simulating Realistic Scenes}\label{sec:scene}
\vspace{-2 mm}
To further bring real-life challenges to the dataset, we added two new features to the models: simulating moving objects as in daily life and simulating variations in lighting.

\vspace{-2 mm}
\subsection{Simulating Scene Change Over Time}\label{sec:rearrange}
\vspace{-2 mm}

As one of the main contributions, we provide certain level of flexibility to automatically configure the arrangement of the furniture objects in order to simulate the traces of daily life. Each object in our database is assigned a convex collision hull, as well as a mass (depending on different objects, in range from 0.05 to 43.3 kg) and a friction coefficient (depending on different surface materials, in range from 0.08 to 0.27) provided by the manufacturers. All those objects are then manually labeled as either \textit{Movable} or \textit{Unmovable}. The former may contain some objects moved on a daily basis, such as mugs, chairs, decorations and shoes etc. The latter may be some big furnitures barely moved, such as fridge, TV, bed and bookshelf etc.
We further use an open-source physics engine, Project Chrono~\cite{chrono}
to dynamically rearrange the furniture objects within a layout. For each configuration, we randomly select 5\% to 45\% movable objects within a layout. We offset the center of gravity slightly to the bottom of the object mesh, in order to avoid tipping over. For each of selected object, we apply a random acceleration (in range from 0.5 to 2 ms$^{-2}$) onto the geometric/mass center of the mesh along a random direction but parallel to the ground. The system takes 10 seconds to allow objects to settle to a physically realistic configuration. In most of the cases, the objects would move less than 2 meter from the original position.

\vspace{-2 mm}
\subsection{Simulating Lighting}
\label{sec:sim_lighting}
\vspace{-2 mm}

\begin{figure}[t]
\centering
\includegraphics[width=1\linewidth]{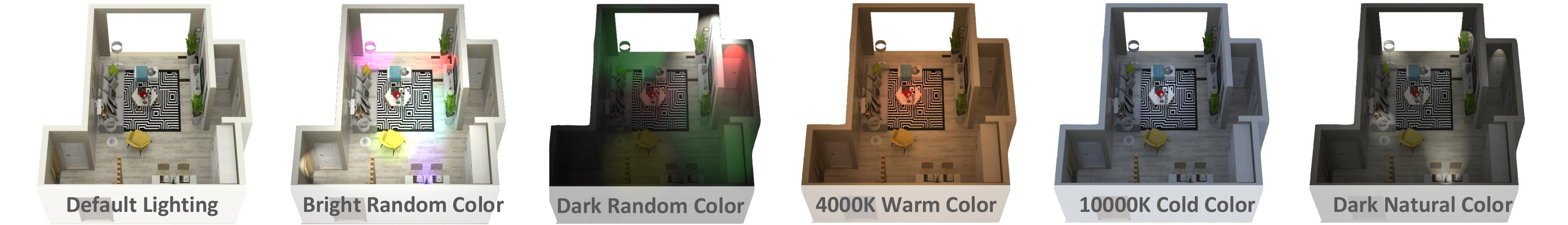}
\vspace{-3mm}
\caption{Lighting setup: natural and random color collection, brightness and temperature.}
\label{fig:lighting}
\vspace{-3mm}
\end{figure}

As another main contribution, we provide a flexibility to manipulate the lighting. A default lighting setup (\textit{light.xml}) is provided for each layout. Those lighting setups include information of lighting type (SunLight, SpotLight and AreaLight), energy, maximum distance (like \textit{PBRT}-v3~\cite{pharr2016physically}), position, direction and brightness, which are provided by manufacturers and have been aligned and fixed to specific objects with lighting sources e.g.\ lamps. Our simulator is able to assist configuring actual RGB value, temperature and brightness of a specific light, or turning it on/off. 
We also support automatically generating different combinations of lighting setups (Fig.~\ref{fig:lighting}).

\vspace{-2 mm}
\section{Trajectory Generation}
\label{sec:traj}
\vspace{-2 mm}

Our dataset provides random trajectories through the scenes. We generated three different types of trajectories, (\textbf{type-1}): two-body random~\cite{Mccormac:ICCV:2017} with height constraints; (\textbf{type-2}): hand-held; and (\textbf{type-3}): look-forward. For each one we have two parameters which can be selected to change the overall speed and angular velocity of the motion. To prevent overly smooth unrealistic trajectories, we augment each of these base trajectories with a learned style model which produces realistic appearing camera jitter.

\vspace{-2 mm}
\subsection{Base Random Trajectories}
\vspace{-2 mm}
Our random trajectories follow the basic two-body random trajectory system described in~\cite{Mccormac:ICCV:2017}.  It simulates two physical bodies randomly moving around and colliding within the indoor free-space.  One defines the camera position, and the other the look-at point. To produce more realistic views (as from a person), the minimum and maximum height is limited to be between 1 m and 2 m.  We also used a slightly randomized up direction with a range from 0$^{\circ}$ to 5$^{\circ}$ difference out of gravity direction.
We improve upon this model by creating two additional modes of trajectory generation.

The first one is designed to simulate a bias in hand-held cameras that tend to look downwards, by restricting the look-at point to be lower than the camera position. The second is designed to provide translational motion along the view direction, a motion type not frequently encountered in the basic two-body system. To achieve this, and give a smooth trajectory, a target point is set in the direction of the velocity from the camera motion, and a proportional force is applied to the current look-at point to smoothly correct the displacement to the current velocity vector.  We also bias the random forces to be in the hemisphere of the velocity vector, to create a smoother look-at target.
We also systematically varied the trajectory parameters.  Unlike in~\cite{Mccormac:ICCV:2017}, where a single set of parameters governed every trajectory, we picked random uniform velocity multiplier from 0.5-5.0 and a random angular velocity scalar from 0.5-3.0 for each trajectory. This feature provides more variety amongst the trajectories.

\vspace{-2 mm}
\subsection{Realistic Trajectories}
\vspace{-2 mm}
An important but neglected factor in existing synthetic datasets for scene understanding is the realism of the trajectory models. Previous work on automated trajectories have used simple and smooth trajectories \cite{Mccormac:ICCV:2017}. Although these trajectories adequately explore the extents of the scene, they do not capture the nuances of real camera motion which can have a significant effect on the visual data. For example, the repetitive impulses in acceleration incurred by the steps of a walking cameraman inevitably adds a significant amount of motion blur to the sequence -- especially in low light indoor conditions. In other papers \cite{Handa:ICRA:2016}, motion-capture trajectories have been captured to solve this -- but such an approach is limited to small datasets where it is feasible to manually capture such trajectories. In this paper, we therefore use a data-driven method to automatically synthesize millions of realistic camera trajectories which satisfy the constraints of the scenes. In particular we harness the recently proposed \textit{Wavenet} architecture~\cite{oord2016wavenet}. The model takes as input a window of previous velocities and outputs the next velocity -- in this manner, synthesizing a trajectory of arbitrary length. 
To ensure that the synthesized trajectory does not collide with objects in the scene, we apply the force model from the preceding Section to the velocities to avoid collisions, and to constrain the overall motion to that of the base random trajectory. We trained the generative model on trajectories from \cite{song2016ssc} and \cite{Sturm_IROS_2012} and synthesized trajectories on the fly during rendering.

\vspace{-2 mm}
\section{\textit{ExaRenderer}: Photo-realistic RGB Renderer}
\label{sec:renderer}
\vspace{-2 mm}
We have developed a fast photo-realistic renderer on top of \textit{Embree}~\cite{Wald:TOG:2014}, an open-source collection of ray-tracing kernels for x86 CPUs. We also have an equivalent variation supporting GPU acceleration to make  full use of our GPU and CPU clusters. Our renderer, coined \textit{ExaRenderer}, supports a common subset of \textit{Path Tracing} operations of leading commercial renderers, and provides flexible APIs to extensively customize the workflow.

\vspace{-2 mm}
\subsection{Path Tracing}
\vspace{-2 mm}
We use the well-known \textit{Path Tracing}~\cite{purcell2002ray} approach to deliver high quality image rendering. Path tracing is a \textit{Monte Carlo} method that approximates realistic Global Illumination (\textit{GI}) for rendering images. It simulates many real-world effects such as soft shadows, depth of field, motion blur and indirect lighting. 
We also support common color bleeding from diffuse surfaces and caustics. Comparing to other popular \textit{GI} approaches, e.g.\ \textit{Photon Mapping}~\cite{jensen2002practical}, \textit{Path Tracing} provides a more realistic caustic effect, requires less memory footprint for large-scale rendering, and is efficient to support dynamic scenes.

\vspace{-2 mm}
\subsection{Sensors Simulation: RGB-D Camera, IMU and Event Camera}
\vspace{-2 mm}

Our renderer supports the pinhole camera model and several lens models such as perspective, depth of field, fisheye and panorama. We used a fixed resolution at 640$\times$480 pixels, and a fixed focus lens at 600 pixels for all the RGB images. We used 5,000$\times$2,500 resolution for panorama and 600$\times$600 for fisheye images. Our renderer simulates camera motion blur by following the method from~\cite{Mccormac:ICCV:2017}; the system takes into account all incoming rays throughout the shutter opening time interval, then integrates the irradiance during rendering.

For inertial measurements,
we fit a cubic B-spline to given control poses (described above), to represent position and orientation, the latter parameterized as a rotation vector. The continuous-time representation lends itself ideally to obtain IMU readings by computing time derivatives at an arbitrary sample rate. Specifically, position (expressed in the coordinates of the static world) is differentiated twice w.r.t.\ time, then acceleration due to gravity is added, and finally the vector is rotated into the frame of reference of the IMU to generate ground truth accelerometer measurements; ground truth rotation rate measurements are obtained from the rotation vector time derivative, where the relationship between the two is given e.g.\ in~\cite{kinder}.
We provide IMU readings at 800 Hz (ground truth, or optionally noisy).

Our rendering framework can also be used to generate output of novel camera designs such as event cameras, e.g.~\cite{lichtsteiner2008128}. We can achieve this by rendering at lower resolution but very high frame rate; then, we assess per-pixel brightness change over time and output events with interpolated timestamps whenever the user-settable intensity threshold is crossed.

Our dataset contains around 20M images which required significant computational power for rendering. To achieve this, our renderer is implemented to be compatible for both CPU and GPU platforms, as well as to support dynamic render distribution onto multiple servers. In this context, we rendered the images on a cluster of 1,300 Nvidia GTX TitanX GPUs and 2,000 Intel Xeon Phi CPUs for around 4 days. We applied 256 samples per pixel (\textit{SPP}) -- the most important parameter to trade-off between image quality and rendering speed.

\vspace{-2 mm}
\subsection{\textit{ViSim}: An Interactive Simulator}
\label{sec:visim}
\vspace{-2 mm}
We implemented an interactive simulator, coined \textit{ViSim}, (Fig.~\ref{fig:teaser}~(D)) to assist creating monocular or stereo camera trajectories, ground truth IMU readings and events given a layout from our database or from any other 3D scenes with \textit{obj} format e.g.\ SUNCG~\cite{song2016ssc} and SceneNet RGB-D~\cite{Mccormac:ICCV:2017} etc. The simulator provides a user-friendly interface while adding our simulation functionality behind. It conntains a \textit{Design View} and a collection of \textit{Parameter Options} to configure camera calibration, distortion coefficients, stereo view, frame rate and travel time, etc. We also support exporting the camera trajectory into other formats e.g.\ SLAMBench~\cite{SLAMBench2015}, SLAMBench2~\cite{Bodin2018ICRA}, EuRoC~\cite{Burri:IJRR_2016}, Freiburg~\cite{Sturm_IROS_2012} and ICL-NUIM~\cite{Handa_ICRA_2014}, and also renderer formats such as \textit{OppositeRenderer} (SceneNet RGB-D)~\cite{Mccormac:ICCV:2017}.

\vspace{-2 mm}
\section{Evaluation on Simultaneous Localization and Mapping}
\label{sec:evaluation}
\vspace{-2 mm}

To verify the quality of the dataset, we have selected several sequences of images of different trajectory types to run \textit{ORBSLAM2.0}~\cite{ORBSLAM2} and \textit{ElasticFusion}~\cite{Whelan2015RSS}.
The verification test uses the RGB-D mode of \textit{ORBSLAM2.0}, using the default configuration parameters.
For each sequence, the absolute trajectory error (\textit{ATE}) is calculated \cite{Sturm_IROS_2012}.

\begin{figure*}[t]
  \centering
  \includegraphics[width=0.95\linewidth]{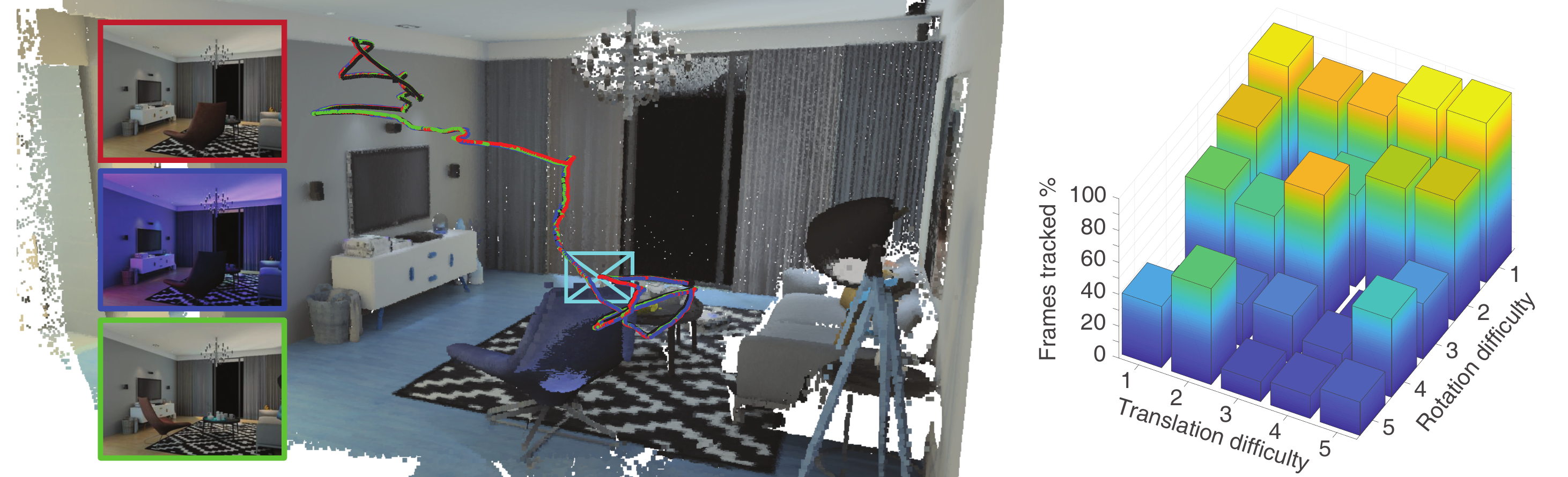}
\caption{ (\textbf{Left}) SLAM under change in lighting and object rearrangement. Estimated trajectories along with the ground truth trajectory are shown overlaid on the dense reconstruction. (\textbf{Right}) Percentage of the frames tracked for 50 trajectories, running \textit{ORBSLAM2.0}.}
\label{fig:slam}
\vspace{-4mm}
\end{figure*}

Fig.~\ref{fig:slam}~(\textbf{Left})
shows the ground truth and estimated trajectories for a randomly chosen type-1 trajectory with a randomly chosen model, in three different scenarios:
a regular scene~(\textbf{Left Top image}),
the same scene with different lighting~(\textbf{Left Middle image}),
and the same scene but with objects displaced~(\textbf{Left Bottom image}).
Trajectory estimation errors for these cases are given in Table~\ref{tb:scene}, rows 1, 2, and 3 respectively. Note that the dense reconstruction is generated by the \textit{ElasticFusion}~\cite{Whelan2015RSS} algorithm.
In the same table, rows 4 and 5 show two other type-1 trajectories with very high position and angular velocities ($v$ and $\omega$ respectively).
\textit{ORBSLAM2.0} is not able to track all frames, indicating the challenge that the trajectories provide.
Rows 5, 6, and 7 show ATE for other types of trajectories with different difficulty levels.

\begin{table}[t]

\begin{minipage}{0.6\columnwidth}
\centering
\begin{tabular}{lll|l|l}
\scriptsize{No} & \scriptsize{Length($m$)} &   \scriptsize{($v$,$\omega$, type)} & \scriptsize{ATE($m$)} & \scriptsize{Description}\\
\hline
\hline
\scriptsize{1} & \scriptsize{21.93} &  \scriptsize{(1,1,1)} & \scriptsize{0.0428} & \scriptsize{a sample model}\\
\scriptsize{2} &  \scriptsize{22.19} &  \scriptsize{(1,1,1)} &  \scriptsize{0.0352}& \scriptsize{.. with different lighting}\\
\scriptsize{3} &  \scriptsize{21.84} &  \scriptsize{(1,1,1)} &  \scriptsize{0.0515}& \scriptsize{.. with objects displaced}\\
\hline
\scriptsize{4} &  \scriptsize{13.88} &  \scriptsize{(9,9,1)} & \scriptsize{0.1701} & \scriptsize{16$\%$ tracked}\\
\scriptsize{5} &  \scriptsize{20.83} &  \scriptsize{(5,6,1)} & \scriptsize{0.0454}&\scriptsize{39$\%$ tracked}\\ 
\hline
\scriptsize{6} &  \scriptsize{17.46} &  \scriptsize{(1,5,1)} & \scriptsize{0.0172}& \scriptsize{type-1}\\ 
\scriptsize{7} &  \scriptsize{22.67} &  \scriptsize{(1,1,2)} & \scriptsize{0.0193}& \scriptsize{type-2}\\ 
\scriptsize{8} &  \scriptsize{4.79} &  \scriptsize{(1,1,3)} & \scriptsize{0.3840}& \scriptsize{type-3, 11$\%$ tracked}
\end{tabular}
\end{minipage}
\hfill
\begin{minipage}{0.39\columnwidth}
\caption{Absolute trajectory error (ATE) for sample sequences when running ORBSLAM2.0. For trajectories, $v$ and $w$ are maximum position and angular velocities in metric units,
with the type explained in Sec.~\ref{sec:traj}:
({type-1}): two-body random; ({type-2}): hand-held; and ({type-3}): look-forward.}
\label{tb:scene}
\end{minipage}
\vspace{-6mm}
\end{table}

Additionally, another random trajectory (length of 24.91~m) was used with several random models. The generated sequences were used to evaluate ORBSLAM2.0. The average ATE was 0.0345~m, with standard deviation of 0.02~m {amongst different scenes}. This indicates that the variation in the models affects the results. To further demonstrate the challenge levels in trajectories, 
Fig.~\ref{fig:slam}~{(\textbf{Right})}
shows the percentage of the tracked frames by \textit{ORBSLAM2.0} for 50 different trajectories with different difficulty levels based on maximum position and angular velocities. As the difficulty level increases, the percentage of the tracked frames drops.

\vspace{-2 mm}
\section{Conclusions}
\label{sec:conclusions}
\vspace{-2 mm}

We have presented a very large synthetic dataset of indoor video sequences that accesses millions of interior design layouts, furniture and object models which were all professionally designed to a highest specification. We then provide variability in terms of lighting and object rearrangement to further devise our scenes and simulate the environment of daily life.
As a result, we obtain highly photo-realistic footage at a high frame-rate.
Furthermore, a large variety of different trajectory types was synthesized, as we believe the temporal aspect should be given closer attention. We demonstrate the usefulness of our dataset by evaluating SLAM algorithms.

In this work, we configured lighting and scene changes in a random fashion due to lack of real-world ground truth for lighting and scene changes. Also, the scene rearrangement was obtained via a physics engine accessing physical parameters e.g. mass, size, friction coefficient etc. Alternatively, a data-driven approach could be used -- which we leave as future work.

\vspace{-2 mm}
\section{Acknowledgements}
\label{sec:acknowledgement}
We would like to thank \url{Kujiale.com} for providing their database of production furniture models and layouts, as well as access to their GPU/CPU clusters. We also thank the Kujiale artists and other professionals for their great efforts into editing and labelling millions of models and scenes. We also highly appreciate the comments and technical support from \textit{Kujiale ExaRendering Group}, as well as helpful discussions and comments from Prof. Andrew Davison and other members of \textit{Robot Vision Group} of Imperial College London. This research is supported by the EPSRC grants PAMELA EP/K008730/1, Aerial ABM EP/N018494/1, and Imperial College London.
\vspace{-2 mm}

\bibliography{egbib}
\end{document}